\pdfoutput=1

\documentclass[11pt]{article}

\usepackage[final]{acl}
\usepackage{natbib}
\usepackage{times}
\usepackage{latexsym}
\usepackage{booktabs}
\usepackage{tabularx}
\usepackage{adjustbox}
\usepackage[T1]{fontenc}

\usepackage{tcolorbox}

\usepackage{amsmath}  
\usepackage{xcolor}   
\usepackage{listings}
\usepackage[utf8]{inputenc}
\usepackage{tcolorbox}
\usepackage{graphicx}
\usepackage{microtype}

\usepackage{inconsolata}

\usepackage{graphicx}

\title{Team Anotheroption at SemEval-2025 Task 8: Bridging the Gap Between Open-Source and Proprietary LLMs in Table QA}

\author{Nikolas Evkarpidi \\
  HSE University \\
  \texttt{nik.evkarpidi@gmail.com} \\\And
  Elena Tutubalina \\
  AIRI \\
  Sber AI \\
  Kazan Federal University \\
  \texttt{ tutubalinaev}@gmail.com \\}

\usepackage{booktabs}    
\usepackage{xcolor}      
\usepackage{array}       

\begin{document}
\maketitle
\begin{abstract}

This paper presents a system developed for SemEval 2025 Task 8: Question Answering (QA) over tabular data. Our approach integrates several key components: text-to-SQL and text-to-code generation modules, a self-correction mechanism, and a retrieval-augmented generation (RAG). Additionally, it includes an end-to-end (E2E) module, all orchestrated by a large language model (LLM). Through ablation studies, we analyzed the effects of different parts of our pipeline and identified the challenges that are still present in this field. During the evaluation phase of the competition, our solution achieved an accuracy of 80\%, resulting in a top-13 ranking among the 38 participating teams. Our pipeline demonstrates a significant improvement in accuracy for open-source models and achieves a performance comparable to proprietary LLMs in QA tasks over tables. The code is available at \href{https://github.com/Nickolas-option/QA_on_Tabular_Data_SemEval2025_Task8}{this GitHub repository}.

\end{abstract}

\section{Introduction}

\begin{figure}[t!]
  \centering
  \includegraphics[height=0.35\textheight, width=0.49\textwidth, keepaspectratio]{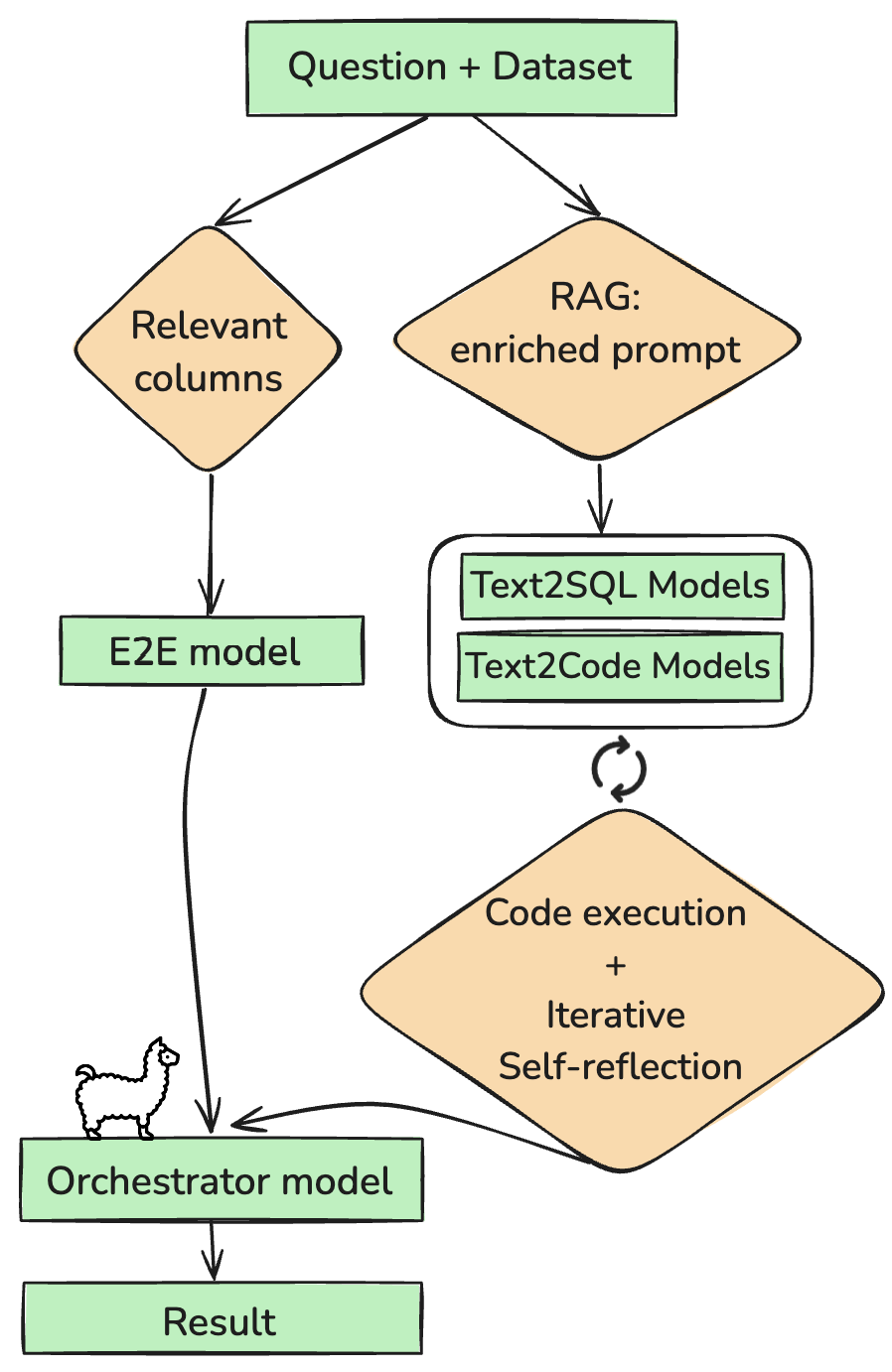}
  \caption{Overview of our system, featuring two solutions: end-to-end (E2E) and code-based. The code-based solution utilizes a self-correction mechanism and retrieval-augmented generation (RAG), with the final decision made by the orchestrator model.}
  \label{fig:overview}
  \vspace{-0.5cm}
\end{figure}

Accessing structured data through natural language (NL) queries is crucial in various fields. However, converting NL into operations that retrieve outputs such as strings, numbers, booleans, or lists continues to be a significant challenge.

This paper outlines our team’s participation in SemEval 2025 Task 8: DataBench \citep{oses-grijalba-etal-2024-question}. This competition assesses question-answering (QA) systems working with tabular data, taking into account various formats, data quality issues, and complex question types. Our aim is to develop a system that accurately retrieves answers from tables, despite challenges such as missing values, inconsistencies, and ambiguous queries. 

We focus on improving Large Language Models (LLMs) using Chain-of-Thought (CoT) reasoning \citep{wang2023planandsolvepromptingimprovingzeroshot, cui2024theoreticalunderstandingchainofthoughtcoherent}. By integrating reasoning-inducing prompts, we enhance LLM-based code generation and decision-making. Additionally, our approach includes an end-to-end (E2E) pipeline and an LLM orchestrator to improve accuracy. To further refine performance, we implement several techniques aimed at reducing model forgetfulness. We introduce structured checklists to help the model verify each step, reducing errors in multi-hop reasoning. By combining prompt engineering, structured reasoning, and workflow optimizations, our system aims to achieve high exact-match accuracy. This paper details our system’s architecture and key techniques (\hyperref[sec:system]{Sec.~\ref{sec:system}}), dataset overview (\hyperref[sec:data]{Sec.~\ref{sec:data}}), and experimental results (\hyperref[sec:results]{Sec.~\ref{sec:results}}).

Our findings offer key insights for enhancing LLM-driven QA for structured data, bridging the gap between open-source and proprietary models. Notably, our development set results showed that open-source LLMs achieved an accuracy of 88\%, surpassing GPT-4o's 74\%.


\section{Related work}
\addcontentsline{toc}{section}{Related Work}
Question Answering (QA) over tabular data has gained significant attention due to the growing need for structured information retrieval \citep{sui2024tablemeetsllmlarge, liu2023rethinkingtabulardataunderstanding, singh2023embeddingstabulardatasurvey, ruan2024languagemodelingtabulardata, r2024navigatingtabulardatasynthesis}. Research in this field has progressed with key datasets, such as FeTaQA \citep{Nan2021FeTaQAFT} and ChartQA \citep{Masry2022ChartQAAB}, as well as a large Wikipedia-based dataset, OpenWikiTable \citep{kweon2023openwikitabledatasetopendomain}. Various methodologies have been explored in recent surveys \citep{fang2024large, Jin2022ASO}, including reinforcement learning and selective classification for text-to-SQL \citep{zhong2017seq2sqlgeneratingstructuredqueries,somov-etal-2024-airi,somov2025}, pre-trained deep learning models \citep{abraham2022tablequeryqueryingtabulardata, Mouravieff2024TrainingTQ}, and few-shot prompting techniques \citep{Guan2024MFORTQAMF}. Building on previous work, we introduce a hybrid LLM-based pipeline that combines multiple techniques to improve performance.    

\section{System Description}\label{sec:system}

Our system, as shown in \hyperref[fig:overview]{Fig.~\ref{fig:overview}}, leverages LLMs and consists of the following key elements:  
\begin{enumerate}
    \item Text-to-SQL and Text-to-Code models to translate NL questions into code executable against tabular data (\hyperref[sec:sql-code-gen]{Sec.~\ref{sec:sql-code-gen}}; \hyperref[sec:pandas-code-gen]{Sec.~\ref{sec:pandas-code-gen}})  
    \item RAG used to enrich prompts with relevant rows and delete irrelevant columns (\hyperref[sec:rag]{Sec.~\ref{sec:rag}});
    \item  Self-correction mechanism used to correct potential errors during execution (\hyperref[sec:self-correction]{Sec.~\ref{sec:self-correction}});
    \item  An E2E answering model to answer questions that target semantic understanding (\hyperref[sec:e2e]{Sec.~\ref{sec:e2e}});
    \item  An orchestrator model to make the final decision between provided solutions (\hyperref[sec:orchestrator]{Sec.~\ref{sec:orchestrator}}).
\end{enumerate}

\begin{figure*}[t]
  \centering
  \includegraphics[width=\textwidth, keepaspectratio]{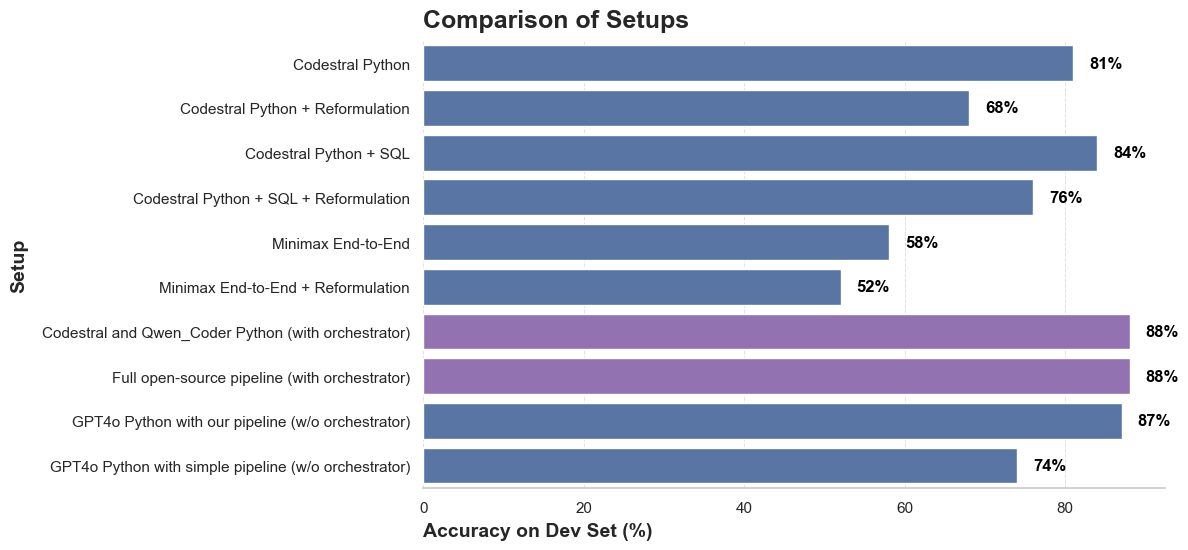}
  \caption{System performance on the dev set. Llama3.3-70b-instruct is used as an orchestrator. As a full pipeline, we're using: Codestral and Qwen Coder for Python and SQL, with Minimax E2E, managed through an orchestrator.}
  \label{fig:experiments}
\end{figure*}

\subsection{Models}
We used state-of-the-art instruction-tuned models featuring various model families: 
Llama \citep{grattafiori2024llama3herdmodels} (version 3.3 with 70b parameters as an orchestrator and 3.2 version with 3b for retrieval),
Codestral (20.51 version) \citep{mistral2024codestral}
and Qwen Coder Instruct (2.5 version with 32b parameters) \citep{hui2024qwen2} for SQL and Code generation, also MiniMax-01 \citep{minimax2025minimax01scalingfoundationmodels} for E2E solution. The selection of the models was driven by their outstanding performance in various benchmarks and the fact that they are open-source.

\subsection{SQL code generation \label{sec:sql-code-gen}}
Here the system resorts to generating SQL queries while using a carefully crafted prompt with a few relevant rows injected in it (\hyperref[sec:rag]{Sec.~\ref{sec:rag}}) and with a suggested list of relevant columns to use (\hyperref[sec:predicting-columns]{Sec.~\ref{sec:predicting-columns}}). The query is executed against the in-memory database (SQLite database via the SQLAlchemy package in our case), and the result is formatted and returned as a potential solution.

\subsection{Pandas code generation \label{sec:pandas-code-gen}}
Here we prompt LLM to generate Python code with the use of Pandas library. The code is then executed against a Pandas Dataframe within a sandboxed environment with a timeout to prevent indefinite loops.  The result of the execution is recorded as a potential solution. Along with the result, we record query success status and error text (if present) for possible future error correction (\hyperref[sec:self-correction]{Sec.~\ref{sec:self-correction}}).

\subsection{Retrieval \label{sec:rag}}

The Databench dataset contains data similar to what you might find in the real world, which presents certain challenges. One of these challenges is accurately filtering data based on specific properties, which often requires contextual knowledge. For example, to answer the question ``How many customers are from Japan?'', the model needs to know that ``japan'' is spelled in lowercase in the dataset. 
To tackle this challenge, we implemented a retrieval step \citep{gao2024retrievalaugmentedgenerationlargelanguage}. We first created sentence embeddings for each relevant column (previously identified in \hyperref[sec:predicting-columns]{Sec.~\ref{sec:predicting-columns}})
 and stored them for efficient searching. When a question was asked, we searched these embeddings to find the top three rows that were most semantically similar. The retrieved data was then used to enrich the LLM's context, enabling the model to answer such questions more accurately and efficiently.

\subsection{Self-correction \label{sec:self-correction}}

The system incorporates a self-correction mechanism \citep{deng2025reforcetexttosqlagentselfrefinement} that attempts to refine solutions that have failed execution attempts. After the failure of the Pandas solution, the system passes meta-info (schema, error message), along with the question back to LLM. Then new solution is generated. The same rules apply to SQL solutions.

\subsection{Reasoning step}
Recent advancements in prompting techniques, such as chain-of-thought prompting ~\citep{wei2022chain}, have demonstrated that large language models (LLMs) can be guided to perform complex reasoning by structuring prompts to include intermediate reasoning steps. In our approach, we leverage this technique by explicitly instructing the LLM to reason extensively before providing its final answer. To extract only the relevant answer from the model's response, we employ fuzzy matching, which allows us to identify and isolate the desired output even when the response contains additional explanatory text or reasoning steps.

\subsection{E2E answer generation \label{sec:e2e}}
This method completely skips the code generation. The dataset is converted into human-readable text (markdown), then given to the LLM along with the underlying question. The model generates a direct answer. Model must put solution into one of the following data formats: Boolean, List, Number or String. The method is used to take advantage of LLM's ability to understand text and, therefore, answer questions about text data from the dataset. 

Unlike code-generation, an E2E solution may only work well in a limited context: that is why we use the Retrieval step (see \hyperref[sec:rag]{Sec.~\ref{sec:rag}})  
 and combine E2E with code-generation approaches to further increase performance.

\subsection{Orchestrator \label{sec:orchestrator}}
We use Llama (3.3 instruct version with 70b parameters) to choose the most probable solution among all presented. The model is provided with several solutions that were successfully executed. Each solution has code and a text-formatted result. Prompt (See \hyperref[lst:orchestrator-prompt]{Appendix prompt})  
 has specific recommendations on how to choose the most probable solution. The model is incapable of generating new solution on the fly and only chooses between the presented options. Further orchestrator's performance analysis is in \hyperref[sec:orchestrator_performance]{Sec.~\ref{sec:orchestrator_performance}}.

\section{Dataset}\label{sec:data}
The dataset comprises 65 publicly available tables across five domains: Health, Business, Social Networks \& Surveys, Sports \& Entertainment, and Travel \& Locations. It retains real-world noise to enhance robustness and includes \textbf{1300} manually curated QA pairs \textbf{in English}, with 500 used for the test set across five answer types: boolean, category, number, list[category], and list[number]. DataBench is provided in two versions: the full dataset and DataBench lite, a smaller subset containing the first 20 rows per dataset. Key dataset statistics are summarized in \hyperref[tab:dataset_stats]{Tab.~\ref{tab:dataset_stats}}. To illustrate dataset diversity, \hyperref[figure:questions-formats]{Fig.~\ref{figure:questions-formats}} presents five representative question types.
Our \textbf{dev set} consists of the first 100 QA pairs, designated for hypothesis testing. 

A notable challenge was handling emojis in column names and textual data, as exact answer matching was required per competition rules, but LLMs struggle with emojis \citep{qiu2024semanticspreservingemojirecommendation}. They often insert spaces or omit them, leading to inaccuracies. 
We mitigated this issue by: 

a) Replacing emojis in column names with unique symbols (hashes) for easier query generation. 

b) Restricting the orchestrator to selecting answers from SQL or Python outputs rather than generating responses, ensuring accuracy.

\subsection{Accuracy Calculation}
The evaluation was conducted using the framework provided in the \href{https://github.com/jorses/databench_eval}{repository}. The evaluation metric was calculated by the rules presented in \hyperref[fig:comparison_rules]{Fig.~\ref{fig:comparison_rules}}. The approach is flexible and provides a fair metric calculation for different pipelines.

\begin{tcolorbox}[colback=blue!10!white, width=0.5\textwidth, 
    boxrule=0.8mm, arc=3mm, top=2mm, bottom=2mm, coltitle=white, fonttitle=\bfseries, 
    title=Comparison Rules, center title]
    \label{fig:comparison_rules}

    \textbf{Numbers}: Truncated to two decimals.

    \vspace{5pt} \hrule \vspace{5pt}

    \textbf{Categories}: Compared directly as-is.

    \vspace{5pt} \hrule \vspace{5pt}

    \textbf{Lists}: Order is ignored.

\end{tcolorbox}

\section{Experiments and Evaluation}\label{sec:results}

This section details the experiments conducted to evaluate our system's performance in Table QA. 
More experiments are in  \hyperref[sec:checklists]{Appx.~\ref{sec:checklists}} and \hyperref[sec:few-shot]{\ref{sec:few-shot}}.

As shown in \hyperref[fig:experiments]{Fig.~\ref{fig:experiments}}, reformulating the question generally decreases accuracy (we discuss why in \hyperref[sec:reformulation]{Sec.~\ref{sec:reformulation}}), as seen in setups like ``Codetral Python + Reformulation'' (68\%) and ``Minimax End-to-End + Reformulation'' (52\%), both of which perform worse than their non-reformulated counterparts. Whereas, adding SQL capabilities tends to increase accuracy, with ``Codetral Python + SQL'' reaching 84\%. The highest accuracy is achieved when multiple models are combined and orchestrated, such as ``Codetral and Qwen Coder Python (with orchestrator)'' and ``Full open-source pipeline (with orchestrator)'', both achieving 88\%, surpassing single-model approaches.

\subsection{Performance of Code-Generation}

Text-to-SQL and text-to-code generation performed well on structured queries but struggled with ambiguous questions that lack explicit context. The self-correction mechanism improved accuracy by refining failed queries. However, unclear queries, such as ``Provide the median number of claims for B2 and S1 kinds'' could lead to misinterpretations, whether computing a single median or separate medians, resulting in incorrect outputs.

\subsection{Effectiveness of E2E Processing}
The E2E approach, which skips code generation and directly answers questions using a textual representation of the table, performed well on questions requiring semantic understanding. For instance, it excelled at answering non-exact questions like ``Is there a patent related to 'communication' in the title?''.
Furthermore, models were given the task of answering the question: ``How many distinct male participants took part in the competition?'' based solely on participants' names. This required the models to infer the participants' sex from their names that is  a task that LLMs typically excel at. However, solving this problem using SQL or Python alone would be quite challenging. In such cases, both systems complemented each other, leveraging the strengths of LLMs for context understanding and the structured data processing power of SQL and Python.

\subsection{Orchestrator Performance \label{sec:orchestrator_performance}}
The orchestrator model, which selects the most probable solution from multiple candidates, generally performed well. However, its accuracy depended heavily on the quality of the candidate solutions. If all candidates were incorrect, the model couldn't generate a correct answer on its own. Additionally, when the majority of candidate answers were incorrect, the orchestrator sometimes failed to select the correct solution, instead favoring the most frequent or popular response.

We propose, for future research, exploring automatic methods to determine whether a given question is better suited for SQL or Python-based querying. This could help the orchestrator make more informed decisions, leading to improved accuracy and efficiency in selecting the correct answer.

\subsection{Question Reformulation \label{sec:reformulation}}
Handling ambiguous or under-specified queries is a key challenge in structured data QA with LLMs \citep{zhao2024icouldveaskedthat}. We tested LLM-based question reformulation to make queries more explicit, but it proved counterproductive (as seen in \hyperref[fig:experiments]{Fig.~\ref{fig:experiments}}). Errors in reformulation at the pipeline's start led to failures without any recovery mechanism. Conversely, without reformulation, some models inferred intent correctly, enabling the orchestrator to choose the right response even if others failed. While reformulation may be less effective with multiple models (e.g., in our case: two SQL, two Python, one E2E), further research is needed to confirm this.

\subsection{Predicting Useful Columns \label{sec:predicting-columns}}

E2E models often encounter challenges when working with long-context data, a difficulty sometimes referred to as ``The Needle In a Haystack problem'' \citep{laban2024summaryhaystackchallengelongcontext}. To address this problem, we introduced  LLM-driven column selection. 
The separate model is given a description of the query and asked to select the most relevant columns before attempting to generate an answer. This method ensures that only the useful parts of the dataset are provided to the E2E model, reducing context length and minimizing the risk of hallucinations.

\subsection{Column name explanation}
Structured datasets often have ambiguous or abbreviated column names, making LLM comprehension challenging. To address this, we introduced column reformulation. For example, given the table ``078 Fires'' and initial data rows, LLMs effectively generated clearer column names. 
\begin{table}[t!]
    \centering
    \begin{tabular}{@{}p{0.49\linewidth} p{0.49\linewidth}@{}}
        \toprule
        \textbf{Original Name} & \textbf{Renamed Column} \\
        \midrule 
        DMC & Duff Moisture Code \\  
        DC & Drought Code \\  
        ISI & Fire Spread Index \\  
        RH & Relative Humidity \\
        \bottomrule
    \end{tabular}
    \caption{Renaming ambiguous column names for clarity using context and LLM insights.}
    \label{tab:renamed_columns}
\end{table}

\hyperref[tab:renamed_columns]{Tab.~\ref{tab:renamed_columns}}
 highlights the ambiguity of original column names, which were clarified through renaming for better usability. However, this poses challenges, as users may refer to original names, necessitating entity recognition for query adjustments, which is beyond the scope of our study. Misinterpretation is also a risk: abbreviations like DC and DMC have multiple meanings, and even strong models can generate incorrect names (e.g., GPT-4o renamed ISI as Fire Spread Index instead of Initial Spread Index). Further research is needed to refine this strategy for effective QA pipeline integration.


\section{Comparison with proprietary models}
The integration of multiple components showcased the potential of open-source LLMs for solving QA tasks over tabular data.
As \hyperref[fig:metrics]{Fig.~\ref{fig:metrics}}
 illustrates, we compared Codestral against GPT-4o \citep{openai2024gpt4ocard}, both utilizing our pipeline. While GPT-4o outperformed Codestral, the performance gap remained within a reasonable range. However, the best results were achieved by a two-model system, which combined Codestral and Qwen Coder for Python code generation, managed by an orchestrator. This setup reached 88\% accuracy, surpassing GPT-4o. By leveraging an orchestrator to optimize the strengths of multiple models, our approach demonstrates that open-source solutions can achieve accuracy levels comparable to proprietary models. Our pipeline applied to GPT4o (w/o orchestrator) also performs well (87\%), resulting in a noticeable improvement over a simpler pipeline, showing the effectiveness of such an approach even for already strong proprietary models. 

\section{Official results}

We ranked in the top 13 out of 38 teams in the competition’s OpenSource-models-only section \citep{osesgrijalba-etal-2025-semeval-2025}, achieving an accuracy score of 80\% on the Databench evaluation as well as on a lite part of the benchmark. Our official results on Databench part of the task are presented in \hyperref[tab:results]{Tab.~\ref{tab:results}}, showing that we significantly outperformed the baseline by 54 points. The best solution achieved a score of 95.20. 
In the global ranking presented in \hyperref[tab:global_results]{Tab.~\ref{tab:global_results}}, which includes proprietary models, we placed in the top 20 out of 53 teams while exclusively using open-source models.
\begin{table}[t!]
    \centering
    \resizebox{0.49\textwidth}{!}
    {
    \begin{tabular}{cccc}
        \toprule
        \textbf{Rank} & \textbf{Codabench ID} & \textbf{Team} & \textbf{Score} \\
        \midrule
        1  & xiongsishi     & TeleAI           & 95.02 \\
        2  & pbujno         & SRPOL AIS        & 89.66 \\
        \textbf{13} & \textbf{anotheroption}  & \textbf{anotheroption} & \textbf{80.08} \\
           & baseline       & stable-code-3b-GGUF               & 26.00 \\
        \bottomrule
    \end{tabular}
    }
    \caption{Official results among open source models}
    \label{tab:results}
\end{table}

\section{Error Analysis}
\subsection{Orchestrator decisions}
In \hyperref[fig:overview]{Fig.~\ref{fig:overview}}, the distribution of orchestrator decision types is shown. Most cases (63.4\%) involved simple confirmation of consensus among identical outputs ('Agreement'). However, in 36.6\% of scenarios, the orchestrator took a more active role: filtering out logically flawed responses (14.6\%), rejecting answers with mismatched data formats (12.2\%), or resolving conflicts between divergent yet seemingly valid outputs (9.8\%).
\begin{figure}[t!]
  \centering
  \includegraphics[height=0.75\textheight, width=0.49\textwidth, keepaspectratio]{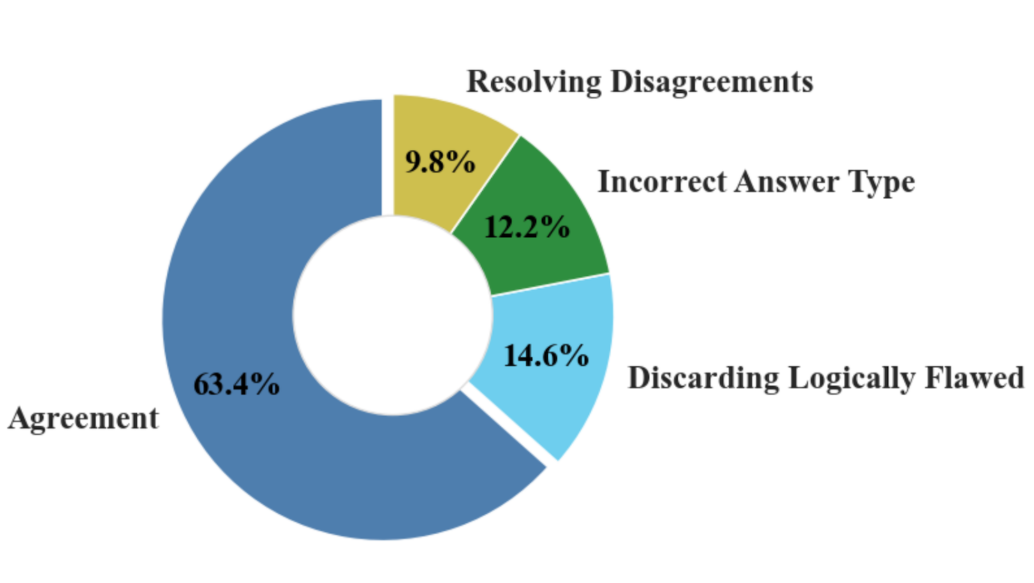}
\caption[Orchestrator Decision Scenarios]{ 
    Distribution of LLM orchestrator decision scenarios (based on 41 questions from the dev set). 
  }
  \label{fig:overview} 
  \vspace{-0.5cm} 
\end{figure}
\subsection{Code-based solution failure analysis}
Some code-based solutions had incorrect syntax. There are several common patterns in which this occurred. 
\begin{itemize}

    \item Incorrect aggregation: queries with broken logical chains, incorrect applications of aggregation functions or ``group by'' operation.

    \item Type unaware operations: the system would often make syntax errors due to incorrect handling, such as trying to retrieve properties over int objects.

    \item Flawed code understanding: errors included attempts to call Pandas methods with incorrect or omitted arguments.
\end{itemize}

And when the code syntactically was correct, there were several common failure patterns.
\begin{itemize}
    \item Subtle logical errors: this often manifests syntactically correct code that nonetheless employs incorrect aggregation, filtering, or sorting logic for the specific dataset.
    \textit{Example: Incorrect identification of the most retweeted author due to flawed aggregation.}
    \item Query misinterpretation: in these cases, the generated code fails to capture the full intent of the query.
    \textit{Example: Returning pokemon name instead of total stats when asked for the ``lowest total stats of pokemon''.}
    \item Data-specific edge cases: generated code struggles with particular data characteristics, such as incorrectly handling null values, emojis, timestamps, or failing to provide a robust approach to tied rankings in sorting or max/min operations.
    \textit{Example: Failure to correctly identify authors of shortest posts due to inaccurate word count.}
\end{itemize}

Identifying these distinct failure types is crucial for improving the overall reliability of the Q\&A system.

\subsection{Self-correction}
The self-correction mechanism was largely ineffective due to the system design involving multiple LLM agents: two for Python and two for SQL. In the vast majority of cases, at least one Python and one SQL agent a runnable solution. As a result, the orchestrator could select a valid answer without having to rely on self-correction.

\section{Conclusion}
We introduced a comprehensive system for QA over tables, showcasing that well-orchestrated open-source models can rival proprietary solutions. We tested various methods: some risked errors, while others improved accuracy and reliability of the system. Our system ranked among the top 13 teams with 80\% accuracy. 
Future work could explore dynamic pipeline selection — automatically determining whether a question requires code-based execution, semantic analysis, or hybrid approaches — to optimize efficiency and accuracy. Additionally, enhancing the orchestrator’s capacity to detect and correct logical inconsistencies in candidate answers could further improve robustness.

\section{Limitations}
The performance of the system exhibits significant variability across different model sizes. Additionally, retrieval systems often encounter challenges when the terms in a query do not align well with the tabular data being searched, and embedding models do not completely address this issue. A notable limitation lies in the generation of candidates for orchestration, where it is possible for all generated responses to be incorrect. This represents a well-known challenge, as identified in prior work \citep{bradley2024llmsmadnesscrowds}, highlighting how certain tasks can prove difficult even for comprehensive groups of large language models (LLMs). In these instances, the system is inherently designed without a mechanism to independently generate a correct answer. Future research could explore potential strategies to address such scenarios effectively.

\section{Ethics Statement}
All models and datasets used are publicly available.
We honor and support the ACL Code of Ethics.

\section*{Acknowledgments}
We acknowledge the computational resources of HPC facilities at the HSE University. The work has been supported by the Russian Science Foundation grant \#23-11-00358.

\bibliography{anthology,custom}

\appendix

\section{Appendix}
\label{appendix:annotations}

\begin{figure*}[t!]
  \centering
  \includegraphics[width=\textwidth, keepaspectratio]{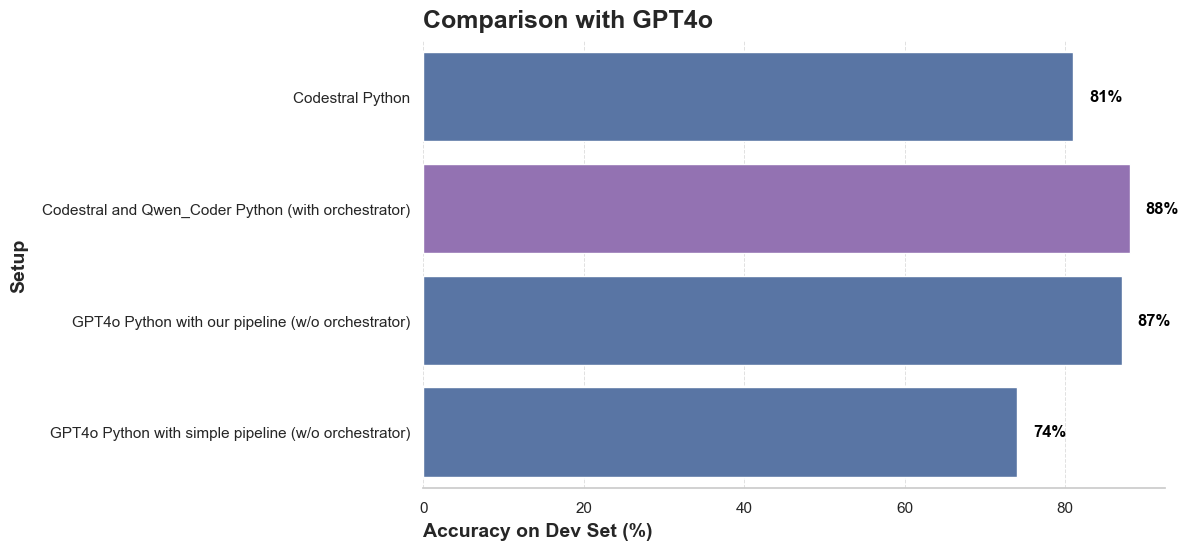}
  \caption{GPT4o comparison with and w/o our pipeline. Llama3.3-70b-instruct used as orchestrator.}
  \label{fig:metrics}
\end{figure*}

\subsection{Checklists and Dialogue-Inducing Prompts} \label{sec:checklists}
During testing, models often skipped crucial instructions, leading to incorrect code generation. To enhance reliability, we implemented checklist-based prompts \citep{cook2024tickingboxesgeneratedchecklists}, enforcing constraints like type matching, entity verification, and logical consistency for more accurate outputs. We also tested dialogue-inducing prompts, where the model simulated a specialist discussion to clarify queries, but this proved superficial, as the model did not actively use the dialogue to correct mistakes.

\subsection{Few-shot Prompting} \label{sec:few-shot}
LLMs perform better with contextual examples ~\citep{liu2021pre}, a phenomenon often referred to as few-shot prompting ~\citep{reynolds2021prompt, brown2020language}. This approach involves providing the model with a small number of task-specific examples before asking it to perform the desired task.  We also experimented with dynamic few-shot prompting \citep{r2024thinksizeadaptiveprompting}, where the model selects relevant examples based on their similarity to the given question. However, this approach requires generating a large number of high-quality examples for each question type, which is both labor-intensive and time-consuming. Additionally, scaling this method could be challenging, as the number of question types may become too large to manage effectively. Due to these limitations, we consider it beyond the scope of our current work.


\begin{table}[h]
    \centering
    \resizebox{0.49\textwidth}{!}{ 
    \begin{tabular}{cccc}
        \toprule
        \textbf{Rank} & \textbf{Codabench ID} & \textbf{Team} & \textbf{Accuracy} \\
        \midrule
        1  & xiongsishi     & TeleAI           & 95.01 \\
        2  & andreasevag    & AILS-NTUA        & 89.85 \\
        \textbf{20} & \textbf{anotheroption}  & \textbf{anotheroption} & \textbf{80.08} \\
           & baseline       & stable-code-3b-GGUF & 26.00 \\
        \bottomrule
    \end{tabular}
    }
    \caption{Results among both open and closed source models}
    \label{tab:global_results}
\end{table}

\begin{table}[h!]
    \centering
    \renewcommand{\arraystretch}{1.1}
    \begin{adjustbox}{max width=\linewidth}
    \begin{tabular}{lp{2.5cm}}
        \toprule
        \textbf{Statistic} & \textbf{Value} \\
        \midrule
        Unique datasets & 49 \\
        Avg. questions per dataset & 20 \\
        Boolean answers (T/F) & 65\% / 35\% \\
        Avg. columns per question & 2.47 \\
        Most common answer types & \\
        \quad Category / Boolean & 199 / 198 \\
        \quad List[Num] / List[Cat] & 198 / 197 \\
        \quad Number & 196 \\
        Columns per dataset (avg./std) & 25.98 / 22.74 \\
        Question length (avg./std) & 61.36 / 18.01 \\
        \bottomrule
    \end{tabular}
    \end{adjustbox}
    \caption{Core statistics illustrating the distribution of questions and answers in DataBench.}
    \label{tab:dataset_stats}
\end{table} 

\begin{figure}[h]
  \centering
  \begin{tcolorbox}[colback=blue!10!white, width=0.4\textwidth, 
      boxrule=0.8mm, arc=3mm, top=2mm, bottom=2mm, coltitle=white, fonttitle=\bfseries,
      title=Data Questions and Formats, center title]
    \textbf{Data Type: Number} \\
    \emph{Q:} What is the average age of our employees? \\
    \emph{Format:} Single numerical value (e.g., 35.2).

    \vspace{5pt} \hrule \vspace{5pt}

    \textbf{Data Type: List[Category]} \\
    \emph{Q:} Unique classifications for employees' education fields? \\
    \emph{Format:} List of categories (e.g., ["Life Sciences", "Marketing"]).

    \vspace{5pt} \hrule \vspace{5pt}

    \textbf{Data Type: List[Number]} \\
    \emph{Q:} Lowest 5 monthly incomes? \\
    \emph{Format:} List of numbers (e.g., [2000, 2100, 2200]).

    \vspace{5pt} \hrule \vspace{5pt}

    \textbf{Data Type: Category} \\
    \emph{Q:} Most common role? \\
    \emph{Format:} Single category (e.g., "Manager").

    \vspace{5pt} \hrule \vspace{5pt}

    \textbf{Data Type: Boolean} \\
    \emph{Q:} Is the highest DailyRate 1499? \\
    \emph{Format:} \texttt{True} or \texttt{False}.
  \end{tcolorbox}
  \caption{Structured data questions and formats.}\label{figure:questions-formats}
\end{figure}


\onecolumn
\clearpage
\newpage

\begin{figure}[h]
    \centering
\begin{tcolorbox}[
    colback=yellow!10, 
    colframe=black, 
    title=prompt for python generation (dialogue)
]

\begin{lstlisting}
1. You are two of the most esteemed Pandas DataScientists engaged in a heated and truth-seeking debate. You are presented with a dataframe and a question. Begin dialogue by rigorously discussing your reasoning step by step, ensuring to address all aspects of the checklist. In your discourse, meticulously articulate the variable type necessary to derive the answer and confirm that each column referenced is indeed present in the dataframe. Conclude your debate by providing the code to answer the question, ensuring that the variable result is explicitly assigned to the answer. Remember, all code must be presented in a single line, with statements separated by semicolons.
2. Refrain from importing any additional libraries beyond pandas and numpy.
3. The dataframe, df, is already populated with data for your analysis; do not initialize it, but focus solely on manipulating df to arrive at the answer.
4. If the question requires multiple entries, always utilize .tolist() to present the results.
5. If the question seeks a single entry, ensure that only one value is output, even if multiple entries meet the criteria.

You MUST FOLLOW THE CHECKLIST, ANSWER EACH OF ITS QUESTIONS (REASONING STEP), AND ONLY THEN OUTPUT THE FINAL ANSWER BASED ON THOSE ANSWERS:
1) How many values should be in the output?
2) Values (or one value) from which column (only one!) should the answer consist of?
3) What should be the type of value in the answer?

Example of a task:
Question: Identify the top 3 departments with the most employees.
<Columns> = ['department', 'employee_id']
<First_row> = ('department': 'HR', 'employee_id': 101)
Reasoning: Count the number of employees in each department, sort, and get the top 3. The result should be a list of department names.
Checklist:
1) The output should consist of 3 values.
2) The values should come from the 'department' column.
3) The type of value in the answer should be a list of strings.
Code: result = df['department'].value_counts().nlargest(3).index.tolist()

Your data to process:
<question> = {question}

- Make absolute sure that all columns used in query are present in the table.
<columns_in_the_table> = {[col for col in df.columns]}
<first_rows_of_table> = {df.head(3).to_string()}
YOUR Reasoning through dialogue and Code (Start final code part by "Code:"):
\end{lstlisting}
\end{tcolorbox}
\caption{Prompt for python generation (dialogue)}
\end{figure}

\begin{figure}[h]
    \centering
\begin{tcolorbox}[
    colback=yellow!10, 
    colframe=black, 
    title=prompt for python generation
]
\begin{lstlisting}
1. You are a best in the field Pandas DataScientist. You are given a dataframe and a question. You should spell out your reasoning step by step and only then provide code to answer the question. In the reasoning state it is essentianl to spell out the answers' variable type that should be sufficient to answer the question. Also spell out that each column used is indeed presented in the table. In the end of your code the variable result must be assigned to the answer to the question. One trick: all code should be in one line separated by ; (semi-columns) but it is no problem for you.
2. Avoid importing any additional libraries than pandas and numpy.
3. All data is already loaded into df dataframe for you, you MUST NOT initialise it, rather present only manipulations on df to calculate the answer.
4. If the question ask for several entries alsways use .tolist().
5. If the question ask for one entry, make sure to output only one, even if multiple qualify.


<...> (same as previous prompt)
\end{lstlisting}
\end{tcolorbox}
\caption{Prompt for python generation (without dialogue)}
\end{figure}

\begin{figure}[h]
    \centering
\begin{tcolorbox}[
    colback=yellow!10, 
    colframe=black, 
    title=prompt for self-correction
]
\begin{lstlisting}
"The following solutions failed for the task: \"{question}\"\n\n"
        + '\n'.join([f'Solution {i+1} Error:\n{traceback}\n' for i, traceback in enumerate(tracebacks)])
        + "\nDF info: \n"
        + "<columns_to_use> = " + str([(col, str(df[col].dtype)) for col in df.columns]) + "\n"
        + "<first_row_of_table> = " + str(df.head(1).to_dict(orient='records')[0]) + "\n"
        + "YOUR answer in a single line of pandas code:\n"
        + "Please craft a new solution considering these tracebacks. Output only fixed solution in one line:\n"
\end{lstlisting}
\end{tcolorbox}
\caption{Prompt for self-correction}
\end{figure}

\begin{figure}[h]
    \centering
\begin{tcolorbox}[
    colback=yellow!10, 
    colframe=black, 
    title=prompt for orchestrator
]
\begin{lstlisting} [label=lst:orchestrator-prompt]
Examples of deducing answer types:
1. If the question is "Do we have respondents who have shifted their voting preference?" the answer type is **Boolean** because the response should be True/False.
2. If the question is "How many respondents participated in the survey?" the answer type is **Integer**
3. If the question is "List the respondents who preferred candidate X?" the answer type is **List** because the response requires a collection of values.
4. If the question is "What is the average age of respondents?" the answer type is **Number** because the response should be a decimal value.
5. If the question is "What is the name of the candidate with the highest votes?" the answer type is **String** because the response is a single textual value.

Given the following solutions and their results for the task: "{question}"

{'   '.join([f'Solution Number {i+1}:  Code: {r["code"]} Answer: {str(r["result"])[:50]} (may be truncated) ' for i, r in enumerate(solutions)])}

Instructions:
- Deduce the most probable and logical result to answer the given question. Then output the number of the chosen answer.
- If you are presented with end-to-end solution, it should not be trusted for numerical questions, but it is okay for other questions.
- Make absolute sure that all columns used in solutions are present in the table. SQL query may use additional double quotes around column names, it's okay, always put them. Real Tables columns are: {df.columns}
- If the column name contain emoji or unicode character make sure to also include it in the column names in the query.
- If several solutions are correct, return the lowest number of the correct solution.
- Otherwise, return the solution number that is most likely correct.
- If the question ask for one entry, make sure to output only one, even if multiple qualify.

You should spell out your reasoning step by step and only then provide code to answer the question. In the reasoning state it is essentianl to spell out the answers' variable type that should be sufficient to answer the question. Also spell out that each column used is indeed presented in the table. The most important part in your reasoning should be dedicated to comparing answers(results) from models and deducing which result is the most likely to be correct, then choose the model having this answer.
First, predict the answer type for the question. Then give your answer which is just number of correct answer with predicted variable type.  Start reasoning part with "REASONING:" and final answer with "ANSWER:".
\end{lstlisting}
\end{tcolorbox}
\caption{Prompt for orchestrator}
\end{figure}

\begin{figure}[h]
    \centering
\begin{tcolorbox}[
    colback=yellow!10, 
    colframe=black, 
    title=prompt for SQL generation
]
\begin{lstlisting}
The task is: {question}
Here are some examples of SQL queries for similar tasks:
Example 1:
Task: Is there any entry where age is greater than 30?
REASONING:
1. Identify the column of interest, which is 'age'.
2. Determine the condition to check, which is 'age > 30'.
3. Use the SELECT statement to retrieve a boolean result indicating the presence of such entries.
4. Apply the WHERE clause to filter rows based on the condition 'age > 30'.
5. Use the EXISTS clause to ensure the query outputs 'True' if any row matches the condition, otherwise 'False'.
6. Verify that the query outputs 'True' or 'False' when presented with a yes or no question.
CODE: ```SELECT CASE WHEN EXISTS(SELECT 1 FROM temp_table WHERE "age" > 30) THEN 'True' ELSE 'False' END;```

Example 2:
Task: Count the number of entries with a salary above 50000.
REASONING:
1. Identify the column of interest, which is 'salary'.
2. Determine the condition to filter the data, which is 'salary > 50000'.
3. Use the SELECT COUNT(*) statement to count the number of rows that meet the condition.
4. Apply the WHERE clause to filter rows based on the condition 'salary > 50000'.
5. Ensure the table name is 'temp_table' and the column name is enclosed in double quotes to handle any spaces or special characters.
CODE: ```SELECT COUNT(*) FROM temp_table WHERE [salary] > 50000;```

Write a correct fault-proof SQL SELECT query that solves this precise task.
Rules:
- Your SQL query should be simple with just SELECT statement, without WITH clauses.
- Your SQL query should output the answer, without a need to make any intermediate calculations after its finish
- Make sure not to use "TOP" operation as it is not presented in SQLite
- If present with YES or NO question, Query MUST return 'True' or 'False'
- If the question asks about several values, your query should return a list
- Equip each string literal into double quotes
- Use COALESCE( ..., 0) to answer with 0 if no rows are found and the question asks for the number of something.

Table name is 'temp_table'.
Available columns and types: {', '.join([f"{col}: {str(type(df[col].iloc[0]))}" for col in column_names])}

Top 3 rows with highest cosine similarity: {get_relevant_rows_by_cosine_similarity(df, question, ai_client).head(3).to_markdown()}
YOUR RESPONSE:
\end{lstlisting}
\end{tcolorbox}
\caption{Prompt for SQL-generation}
\end{figure}

\begin{figure}[h]
    \centering
\begin{tcolorbox}[
    colback=yellow!10, 
    colframe=black, 
    title=prompt for E2E model
]
\begin{lstlisting}
Question: {question}
    Dataset: {dataset_text}

    Analyze the data. Provide your final answer to the question based on the data.
    If the question assumes several answers, use a list. Your answer should be in the form of one of the following:
    1. Boolean (True/False)
    2. List (e.g., ['Tree', 'Stone'])
    3. Number (e.g., 5)
    4. String (e.g., 'Spanish')

    Give extensive reasoning and then fianlly provide the answer starting with string "Final Answer:" in one of the four formats presented above (Boolean, List, Number, String). Your response should then be finished.
\end{lstlisting}
\end{tcolorbox}
\caption{Prompt for E2E model}
\end{figure}

\end{document}